\documentclass[sigconf]{acmart}

\copyrightyear{2025}
\acmYear{2025}
\setcopyright{cc}
\setcctype{by}
\acmConference[SIGIR '25]{Proceedings of the 48th International ACM SIGIR Conference on Research and Development in Information Retrieval}{July 13--18, 2025}{Padua, Italy}
\acmBooktitle{Proceedings of the 48th International ACM SIGIR Conference on Research and Development in Information Retrieval (SIGIR '25), July 13--18, 2025, Padua, Italy}
\acmDOI{10.1145/3726302.3730284}
\acmISBN{979-8-4007-1592-1/2025/07}

\usepackage{multirow}
\usepackage{pifont}
\usepackage{xspace}

\definecolor{framegray}{gray}{0.96}  

\newcommand{\datasetname}{\textsc{WikiHint}\xspace}
\newcommand{\evaluatorname}{\textsc{HintRank}\xspace}
\begin{document}

\title{\datasetname: A Human-Annotated Dataset for Hint Ranking and Generation 
}


\author{Jamshid Mozafari}
\orcid{0000-0003-4850-9239}
\affiliation{%
  \institution{University of Innsbruck}
  \city{Innsbruck}
  \state{Tyrol}
  \country{Austria}}
\email{jamshid.mozafari@uibk.ac.at}

\author{Florian Gerhold}
\affiliation{%
  \institution{University of Innsbruck}
  \city{Innsbruck}
  \state{Tyrol}
  \country{Austria}}
\email{florian.gerhold@student.uibk.ac.at}

\author{Adam Jatowt}
\orcid{0000-0001-7235-0665}
\affiliation{%
  \institution{University of Innsbruck}
  \city{Innsbruck}
  \state{Tyrol}
  \country{Austria}}
\email{adam.jatowt@uibk.ac.at}


\begin{abstract}
The use of Large Language Models (LLMs) has increased significantly with users frequently asking questions to chatbots.
In the time when information is readily accessible, it is crucial to stimulate and preserve human cognitive abilities and maintain strong reasoning skills. This paper addresses such challenges by promoting the use of hints as an alternative or a supplement to direct answers. We first introduce a manually constructed hint dataset, \datasetname, which is based on Wikipedia and includes 5,000 hints created for 1,000 questions. We then finetune open-source LLMs for hint generation in answer-aware and answer-agnostic contexts. We assess the effectiveness of the hints with human participants who answer questions with and without the aid of hints. Additionally, we introduce a lightweight evaluation method, \evaluatorname, to evaluate and rank hints in both answer-aware and answer-agnostic settings. Our findings show that (a) the dataset helps generate more effective hints, (b) including answer information along with questions generally improves the quality of generated hints, and (c) encoder-based models perform better than decoder-based models in hint ranking.
\end{abstract}





\maketitle

\section{Introduction}\label{s:introduction}


\begin{figure}
    \centering
    \includegraphics[width=0.9\columnwidth]{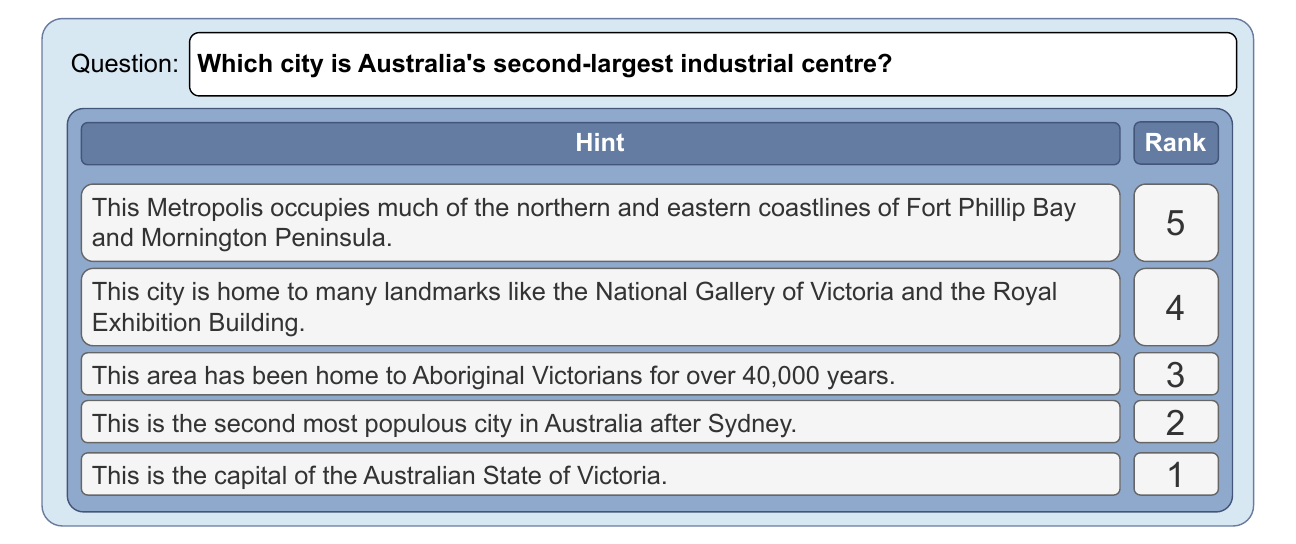}
    \caption{Hints for a question from the \datasetname dataset, with their corresponding rankings (1 being the highest and 5 the lowest) which let users find the answer. Note that the opposite arrangement of the hints would make the answer finding task easiest when hints are read from rank 1 to 5.}
    \label{fig:dataset_sample}
    \Description{}
\end{figure}
\begin{figure*}
	\centering
	\includegraphics[width=0.8\linewidth]{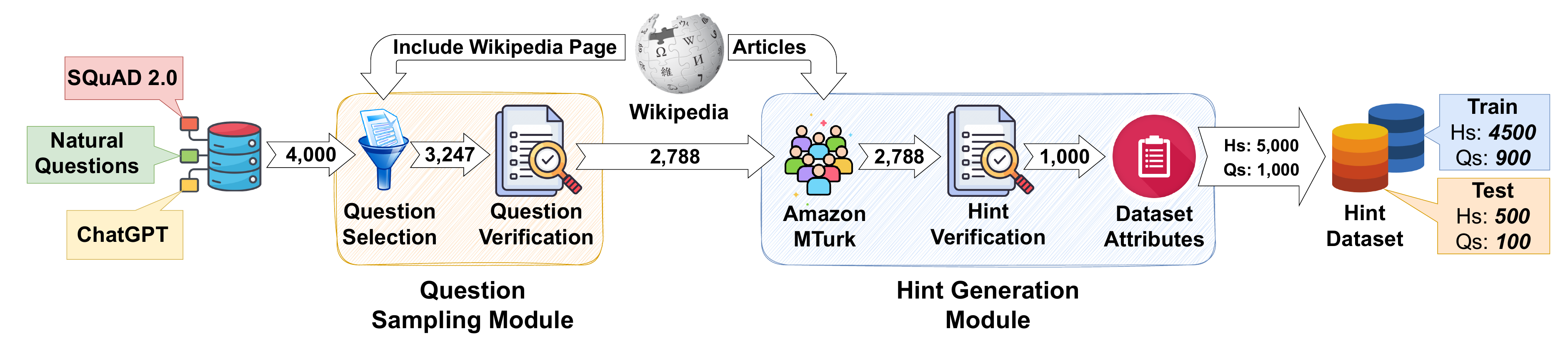}
	\caption{Pipeline of \datasetname dataset generation. The numbers in the arrows indicate the counts of output questions. \emph{Qs} and \emph{Hs} denote the number of questions and hints, respectively.}
	\label{fig:pipeline}
        \Description{}
\end{figure*}

In recent years, question answering (QA) systems have risen in importance, offering users the opportunity to effortlessly receive answers to different types of questions~\cite{DBLP:journals/ftir/MaviJJ24,karpukhin-etal-2020-dense,zhao2020sparta,abdel2023deep}. 
The rapid development of Large Language Models (LLMs)~\cite{2023arXiv231211805G, 2023arXiv230308774O, 2024arXiv240721783D} has mainly contributed to this, impacting also many other Natural Language tasks (NLPs)~\cite{2024arXiv240512819Q}. While the benefits of using LLMs for current information access are quite evident, concerns have been raised in relation to their potential effect on human cognitive development. One such worry relates to the potential weakening of important cognitive skills of users like thinking, reasoning, and remembering due to the expected widespread use of the automatic question answering technologies, in particular, ones backed by powerful AI technologies ~\cite{Heersmink2024}.
Users who will rely mainly on the solutions presented by AI's might also be discouraged to practice and improve their reasoning abilities~\cite{ALFREDO2024100215}.
For example,~\citet{DARVISHI2024104967} demonstrate that students are more likely to depend on AI assistance instead of learning from it. \citet{app14104115} examine the impact of LLMs as an automated problem-solving technology on education and learning outcomes, demonstrating that such systems can negatively affect the development of learning skills. Furthermore, psychological studies confirm the importance of obtaining an answer independently, enhancing user self-confidence~\cite{bandura2013role}. Letting users come up with the correct answers by themselves should then also contribute to the positive psychological effect, potentially increasing their self-confidence~\cite{usher2006sources}. 

While the problem of over-reliance on chatbots is complex and still not researched well, we wish to promote approaches that involve humans in the answer finding process. Providing users with hints rather than direct answers is one such approach.
Hints are designed to offer subtle clues that guide users toward the correct answer without explicitly revealing it~\cite{hume1996hinting}. Unlike the concept of framing (aka "scattered conceptualization") \cite{entman1993framing} which elevates particular pieces of information to promote a certain view or cause bias \cite{garrido2021survey,farber2020multidimensional}, hints help humans to recall previously known information or guide them in the reasoning process.  
In this study, we define hints for factoid questions\footnote{Factoid questions are those with answers in the form of entities such as locations, persons, etc.} as clues that encourage users to think critically, reason, and draw on their knowledge to deduce the answer rather than having it explicitly provided.
This approach aims to support and enhance cognitive skills such as critical thinking, reasoning, and memory. 
To ensure that hints are meaningful and widely accessible, we generate them based on external knowledge and well-established facts. For this purpose, we select Wikipedia as the primary source, as it is one of the most widely recognized and comprehensive repositories of factual information.

In this paper, we propose the first manually curated dataset for the Automatic Hint Generation (HG) task, called \datasetname\footnote{\url{https://github.com/DataScienceUIBK/WikiHint}}, which has been constructed and designed for hint generation, ranking, and evaluation. 
Figure~\ref{fig:dataset_sample} shows a question from the \datasetname dataset, with its hints and the corresponding ranks reflecting the usefulness of each hint.  
We explore the performance of various LLMs in generating hints across different scenarios, including vanilla and finetuned models. We also examine the quality of the generated hints using both answer-aware and answer-agnostic approaches. Finally, we assess the effectiveness of a novel evaluation method for hint ranking, called \evaluatorname, and compare it with other automatic evaluation techniques. Ranking hints allows arranging them either from easiest to hardest which can be used for selecting the most useful hints for a given question, or from hardest to easiest (as shown in Fig. \ref{fig:dataset_sample}). The latter arrangement could find applications in quizzes, knowledge assessment scenarios, or in certain entertainment use cases (e.g., in games).

In summary, we make the following contributions in this paper:
\begin{itemize}
	\item We release the first manually created dataset called \datasetname for the HG task containing 5,000 hints and 1,000 questions. 
	\item We propose an automatic evaluation method for ranking hints called \evaluatorname and compare with other evaluation methods.
	\item We finetune and evaluate diverse LLMs on 
 \datasetname to assess the dataset quality and LLMs' capabilities in hint generation and ranking.
    \item We present several novel observations, including the findings of a positive correlation between hint convergence and helpfulness, an inverse correlation between hint length and helpfulness, and the superiority of the answer-aware approach over the answer-agnostic approach in hint generation.
\end{itemize}

\section{Related Work} \label{s:related_works}
Automatic question answering (QA)~\cite{karpukhin-etal-2020-dense,zhao2020sparta,abdel2023deep} and question generation (QG)~\cite{kurdi2020systematic,lu2021survey,zhang2021review} have progressed a lot in the last years. These tasks have seen numerous different datasets~\cite{trischler2016newsqa, rajpurkar2016squad,zhang2019bertscore}, and evaluation metrics~\cite{nema2018towards,mavi2022survey} proposed. The research related to hint generation is however still scarce despite the fact that hinting is a common mechanism used by humans for question answering, and that automatic hint generation could be regarded as the third missing task alongside QA and QG. The prior research however focused mainly on generating hints for programming~\cite{price2019comparison,kochmar2022automated,barnes2010automatic,mcbroom2021survey}. 

In the context of intelligent tutoring systems, automated,  machine-learning-based methods have been shown to be effective in generating feedback and hints, reducing reliance on expert-crafted rules~\cite{10.1007/978-3-030-52240-7_26, Kochmar2022}. However, the effectiveness of LLMs in pedagogical tasks remains an open challenge. For instance,~\citet{2022.EDM-short-papers.54} investigated the teaching capabilities of LLMs such as Blender~\cite{jiang-etal-2023-llm} and GPT-3~\cite{NEURIPS2020_1457c0d6} in educational dialogues and found that, while these models perform well in conversational uptake, they lack pedagogical effectiveness, particularly in providing helpful feedback. To address this limitation, several studies have proposed different approaches to improving LLM-generated feedback. 
For example,~\citet{kulshreshtha2022few} leveraged a few-shot question generation strategy to create personalized feedback for mathematics problems, demonstrating improved student learning outcomes. Similarly,~\citet{mcnichols2023automated} explored the use of in-context learning and various LLMs for distractor generation in multiple-choice mathematics questions but found that the models still exhibit weaknesses, requiring further research for improvement. Additionally,~\citet{tonga2024automatic} investigated the impact of different LLMs and prompting strategies on the quality of generated feedback and hints, highlighting that the effectiveness of LLM-generated assistance is highly dependent on the choice of model and the prompting approach used.

Automatic hint generation for factoid questions was first addressed by~\citet{jatowt2023automatic}. However, the authors neither released a dataset nor utilized LLMs, focusing instead on hints generated from selected Wikidata\footnote{\url{https://www.wikidata.org/}} predicates. Moreover, their work only considered an answer-aware setting in hint generation and did not explore the hint ranking task.
Subsequently,~\citet{triviahg_mozafari} released the first synthetic dataset for hint generation (HG) called TriviaHG, which was automatically generated using LLMs. However, the automatic generation increases the likelihood of false information within the dataset, particularly due to the hallucination phenomenon of LLMs. The authors also introduced the first automatic evaluation method, called Convergence, for assessing hint quality. However, that method requires substantial computational resources as it relies on LLMs for evaluation. \citet{mozafari-etal-2024-exploring} also demonstrated that hints automatically generated from TriviaHG can serve as a context for QA systems, outperforming alternative approaches such as retrieval-based and generative methods in RAG. Interested readers can also refer to the recent survey of~\citet{2024arXiv240404728J} who discuss various types of hints and challenges associated with hint generation and evaluation.

To our knowledge, \datasetname is the first data set for the HG task with questions and their hints verified by humans. Our proposed automatic evaluation method for hint ranking is also the first for this task that does not rely on LLMs and is lightweight enough to be used locally.

\section{\datasetname Dataset} \label{s:dataset}

The absence of verified high-quality hint datasets poses a significant challenge given the demanding data requirements of LLMs for their effective training. 
In this section, we outline the process for constructing \datasetname dataset.
Figure~\ref{fig:pipeline} provides an overview of the pipeline of the dataset generation process, which we explore in detail in the following sections.

\begin{figure}
	\centering
	\includegraphics[width=\columnwidth]{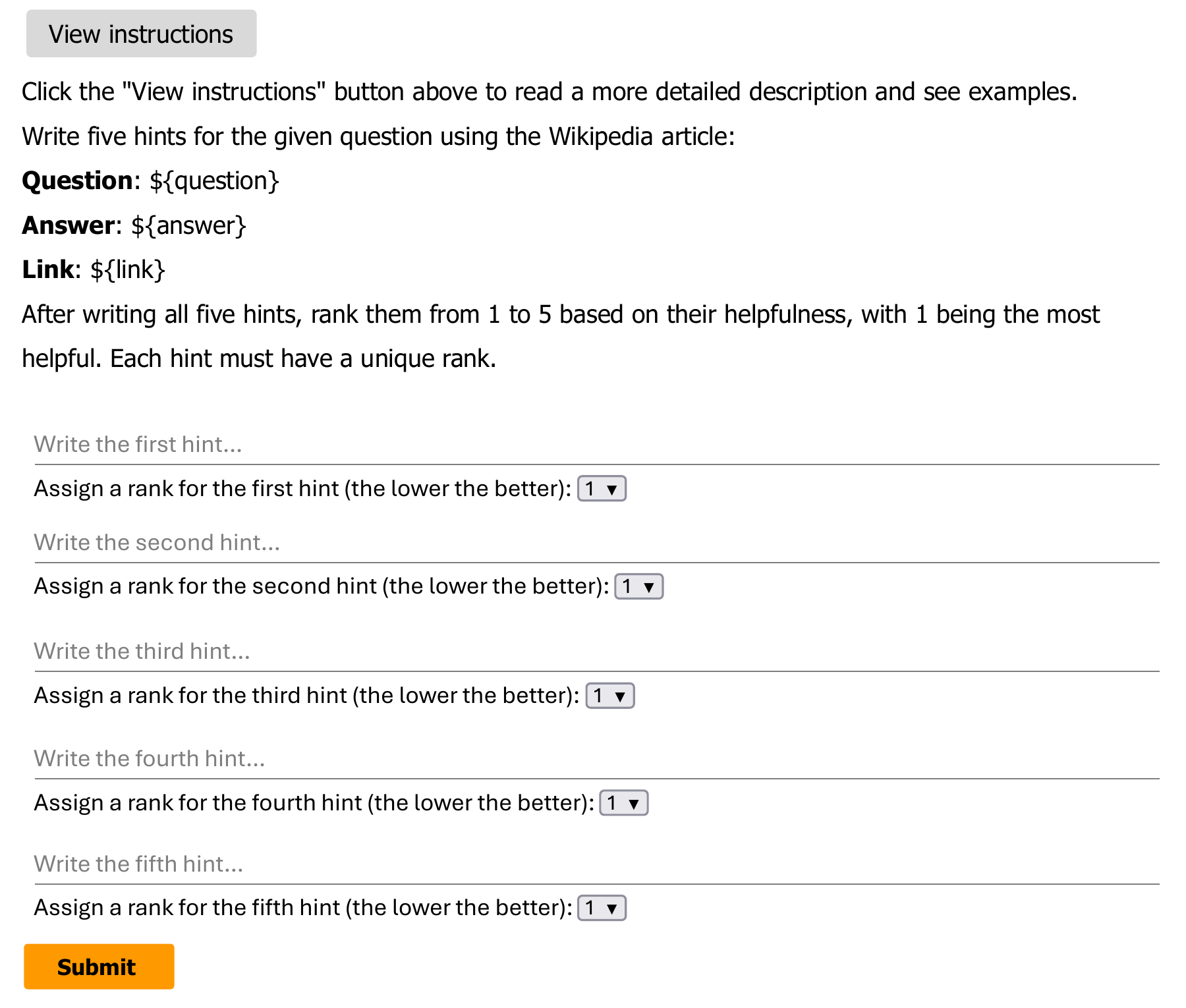}
	\caption{The MTurk Worker interface for generating and ranking hints. \texttt{\$\{question\}} represents the question presented to the worker, \texttt{\$\{answer\}} is the corresponding answer, and \texttt{\$\{link\}} provides the link to the relevant Wikipedia page.}
	\label{fig:mturk_interface}
        \Description{}
\end{figure}

\subsection{Question Sampling Module} \label{ss:question_sampling_module}
We first select questions that meet our criteria for hint generation.
\subsubsection{Question Selection}\label{ss_question_selection}
We incorporated 4,000 questions, including randomly selected questions from popular QA datasets such as SQuAD 2.0~\cite{rajpurkar2018know} and Natural Questions (NQ)~\cite{kwiatkowski2019natural}, as well as AI-generated questions using ChatGPT~\cite{ChatGPT}. The following prompt was used for question generation with ChatGPT:

\begin{center}
\fcolorbox{black!75!black}{framegray}{%
  \begin{minipage}{0.9\linewidth}
    \footnotesize
      \emph{Can you give me 10 questions where the answer is ANSWER? Please put them in a CSV file with answer=ANSWER and link=WIKIPEDIA\_LINK where each question has an answer and link. Make sure to put the questions in quotation marks.}
  \end{minipage}
}
\end{center}

\begin{figure}
	\centering
	\includegraphics[width=\columnwidth]{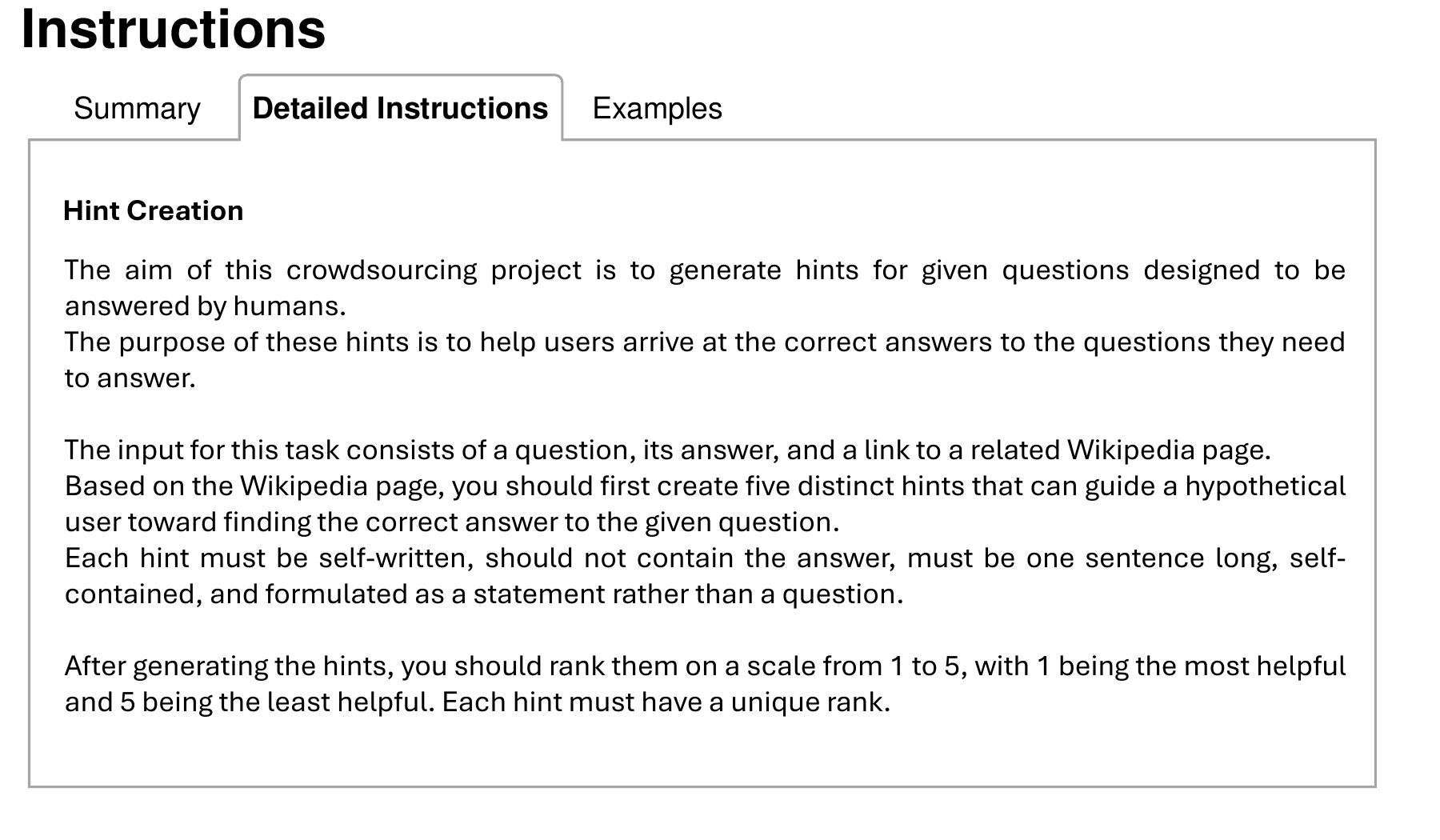}
	\caption{The instructions for the hint generation and ranking tasks on the Amazon MTurk platform.}
	\label{fig:mturk_detailed}
        \Description{}
\end{figure}

\begin{table}
	\centering
    \caption{A detailed description of attributes of a question.}
    \resizebox{\linewidth}{!}{%
	\begin{tabular}{@{}ll@{}}
		\toprule
		\textbf{Attribute}      & \textbf{Description}                                                        \\ \midrule
		question                & The content of the question.                                                \\
		major                   & The major category of the question.                                         \\
		minor                   & The specific sub-category of the question.                                  \\
		entity                  & Content of an entity.                                                        \\
		ent\_type               & Type of entity (e.g., GPE, PERSON).                                         \\
		start\_index            & Start index of the entity in the question.                                  \\
		end\_index              & End index of the entity in the question.                                    \\
		wikipedia\_page\_title  & The title of the Wikipedia page corresponding to the entity in question.    \\
		wiki\_views\_per\_month & Number of views per month of the Wikipedia page for the entity in question. \\
		normalized\_views       & Views normalized to scale from 0 to 1 for the entity in the question.       \\
		readability             & The readability score of the question.                                             \\
		familiarity             & The familiarity score of the question.                                             \\
		difficulty              & The difficulty level of the question.                              \\ \bottomrule
	\end{tabular}%
    }
	\label{tbl:question_attributes}
\end{table}

where \texttt{WIKIPEDIA\_LINK} is the URL of the Wikipedia page corresponding to \texttt{ANSWER}.
As mentioned above, we also selected questions from SQuAD 2.0 and NQ making sure that their answers had dedicated Wikipedia articles with sufficiently long content. After this filtering process, 3,247 questions remained for the next steps.

\subsubsection{Question Verification}\label{ss_q_verification}  
We manually verified the correctness of all the selected questions, discarding those that were too general or contained incorrect answers. Additionally, we ensured that all questions included a question mark, adding it where necessary. We also reviewed all Wikipedia links and corrected any character errors in the URLs. After this filtering process, 2,788 questions remained for the next steps.

\subsection{Hint Generation Module} \label{ss:hint_generation_module}
In this module, we generate hints and rank them.

\subsubsection{Amazon MTurk}\label{ss_amazon_mtruk}
The crowdsourcing platform Amazon Mechanical Turk\footnote{\url{https://www.mturk.com/}} was then used to distribute the hint creation task among multiple workers. The instructions shown to crowdworkers asked them to create five hints for a question and its associated Wikipedia article. After generating the hints, the same workers were asked to rank them on a scale from 1 to 5, with 1 being the most helpful in finding correct answers and 5 being the least helpful. 
Figure~\ref{fig:mturk_interface} displays the annotators' interface for hint generation while Figure~\ref{fig:mturk_detailed} provides detailed instructions to further assist the crowdworkers.


\begin{table}
    \centering
	\caption{A detailed description of attributes of an answer.}
    \resizebox{\linewidth}{!}{%
	\begin{tabular}{@{}ll@{}}
		\toprule
		\textbf{Attribute}      & \textbf{Description}                                                          \\ \midrule
		answer                  & The actual answer.                                                            \\
		entity                  & Content of an entity as identified within the answer.                         \\
		ent\_type               & Type of entity.                                                               \\
		start\_index            & Start index of the entity in the answer.                                      \\
		end\_index              & End index of the entity in the answer.                                        \\
		wikipedia\_page\_title  & The title of the Wikipedia page corresponding to the entity in the answer.    \\
		wiki\_views\_per\_month & Number of views per month of the Wikipedia page for the entity in the answer. \\
		normalized\_views       & Views normalized to scale from 0 to 1 for the entity in the answer.           \\
		familiarity             & The familiarity score of the answer.                                               \\  \bottomrule
	\end{tabular}%
    }
	\label{tbl:answer_attributes}
        \Description{}
\end{table}

\begin{table}
    \centering
    \caption{A detailed description of attributes of a hint.}
    \resizebox{\linewidth}{!}{%
	\begin{tabular}{@{}ll@{}}
		\toprule
		\textbf{Attribute}      & \textbf{Description}                                                                  \\ \midrule
		hint                    & Hint provided for the question.                                            \\
		source                  & URL source of the hint.                                                               \\
		entity                  & Entities mentioned in the hint.                                            \\
		ent\_type               & Category of each entity (e.g., PERSON, GPE) mentioned in the hint.                    \\
		start\_index            & Specific start index where the entity is found in the hint text.                      \\
		end\_index              & Specific end index where the entity is found in the hint text.                        \\
		wikipedia\_page\_title  & The title of the Wikipedia page corresponding to the entities in the hint.            \\
		wiki\_views\_per\_month & Number of views per month of the Wikipedia pages for the entities of the hint. \\
		normalized\_views       & Views normalized to scale from 0 to 1 for the entity in the hint.                     \\
		relevance               & The relevance score.                                                         \\
		readability             & The readability score.                                                       \\
		convergence             & The convergence score.                                                       \\
        familiarity             & The familiarity score.                                                       \\
        answer\_leakage          & The answer leakage score.                                                   \\
		rank                    & Priority or helpfulness rating of the hint.                                            \\ \bottomrule
	\end{tabular}%
    }
	\label{tbl:hint_attributes}
        \Description{}
\end{table}

\subsubsection{Hint Verification}\label{ss_h_verification}
Each data submission was then manually reviewed by three independent judges for quality and either approved or rejected based on their evaluation. The following section outlines the detailed criteria used in the selection process.

\begin{itemize}
        \item The hint must not explicitly include the exact answer.
        \item The hint must be a sentence.
        \item The hint must be specific, not generic.
        \item The hint must be from the corresponding Wikipedia page.
        \item The hint must have a unique rank.
\end{itemize}

The most common reasons for rejection were hints that directly revealed the answers (answer leakage) and hints that were single words instead of complete sentences. Among the 2,788 submissions reviewed, 1,788 were rejected and 1,000 were accepted. 
\subsubsection{Dataset Attributes}\label{ss_dataset_attributes}
We prepared several attributes to include for each question, answer, and hint, as shown in Tables~\ref{tbl:question_attributes}, \ref{tbl:answer_attributes}, and \ref{tbl:hint_attributes}, respectively. We discuss some of them in Section~\ref{s:evaluation_method}.

Two key attributes of a question are \emph{major} and \emph{minor}, which indicate the question type. To extract these attributes, we employ question classifiers, which categorize natural language questions into predefined classes based on their intended meaning. For example, the question \emph{Who is the director of the Godfather movie?} falls into the \textit{PERSON:INDIVIDUAL} class, while \textit{What is the capital of Austria?} belongs to the \textit{LOCATION:CITY} class.  
To develop an effective question classifier, we fine-tune the RoBERTa~\cite{liu2019roberta} model using the TREC Question Classification dataset~\cite{li-roth-2002-learning}, resulting in a model called QT Detector. This model achieves an accuracy of $92.8\%$ on the TREC Question Classification dataset\footnote{The accuracy of the QT Detector on the TREC Question Classification dataset is $92.8\%$}. Using QT Detector, questions are categorized into hierarchical classes with varying levels of granularity, encompassing 5 coarse-grained classes and 50 fine-grained classes.

Finally, we divide the questions and hints into train and test subsets. The train subset includes 4,500 hints for 900 questions, while the test subset contains 500 hints for 100 questions.

\begin{figure}
	\centering
	\includegraphics[width=0.7\columnwidth]{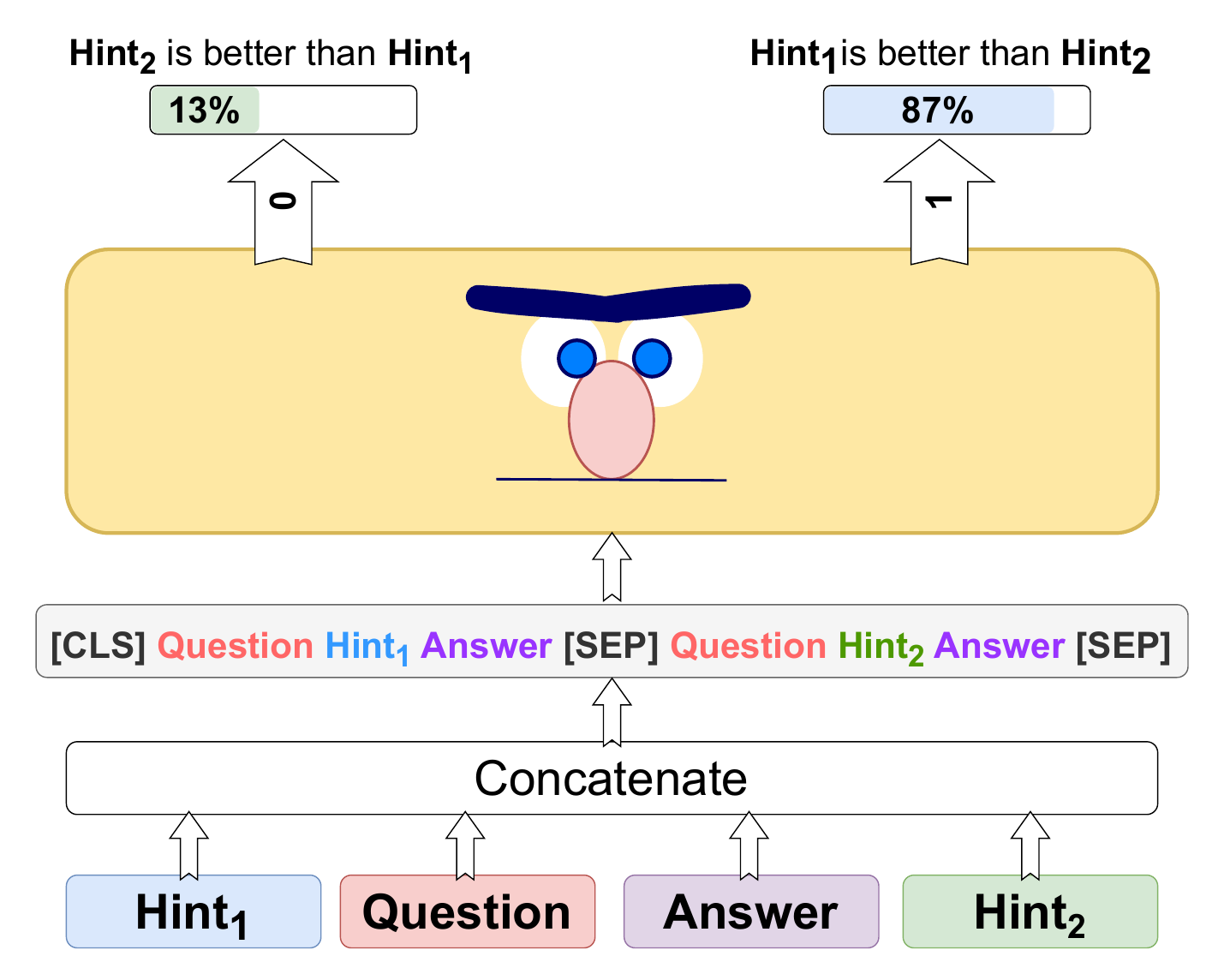}
	\caption{The \evaluatorname method. First, the inputs are concatenated, and special tokens are added to format them for the BERT model. Finally, the BERT model determines which hint is more helpful.}
	\label{fig:evaluation_method}
        \Description{}
\end{figure}

\begin{figure}
	\centering
	\includegraphics[width=0.7\columnwidth]{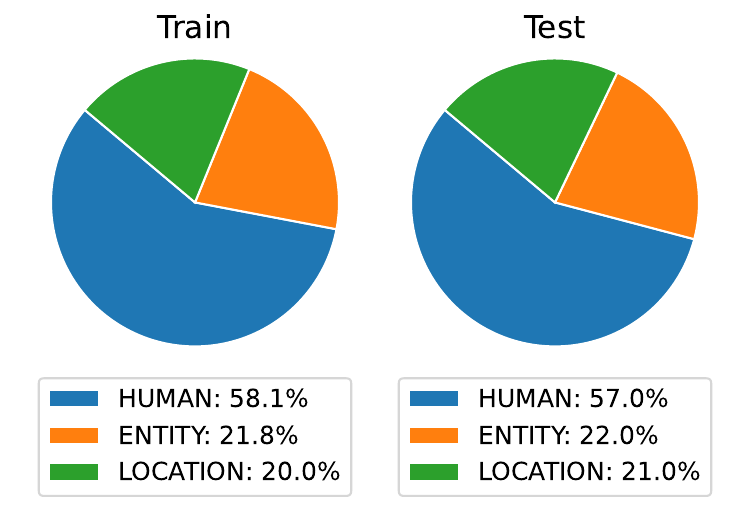}
	\caption{The distribution of the Train and Test subsets.}
	\label{fig:dataset_statistics}
        \Description{}
\end{figure}

\section{Evaluation Approaches} \label{s:evaluation_method}

In this section, we begin by introducing the evaluation metrics used to assess the quality of hints. These metrics are available through the HintEval~\cite{mozafari2025hinteval} framework, which provides a standardized setup for hint evaluation\footnote{\url{https://github.com/DataScienceUIBK/HintEval}}. Following this, we present \evaluatorname, our proposed method for automatic hint ranking.

\subsection{Evaluation Metrics}\label{ss_evaluation_metrics}
\citet{triviahg_mozafari} introduced five evaluation metrics for assessing hint quality: Relevance, Readability, Convergence, Familiarity, and Answer Leakage. While the authors did not propose automatic methods for Relevance, Readability, and Answer Leakage, in this paper, we employ automated approaches for these metrics. In the following, we provide a detailed description of all metrics.

\begin{figure}
	\centering
	\includegraphics[width=0.8\columnwidth]{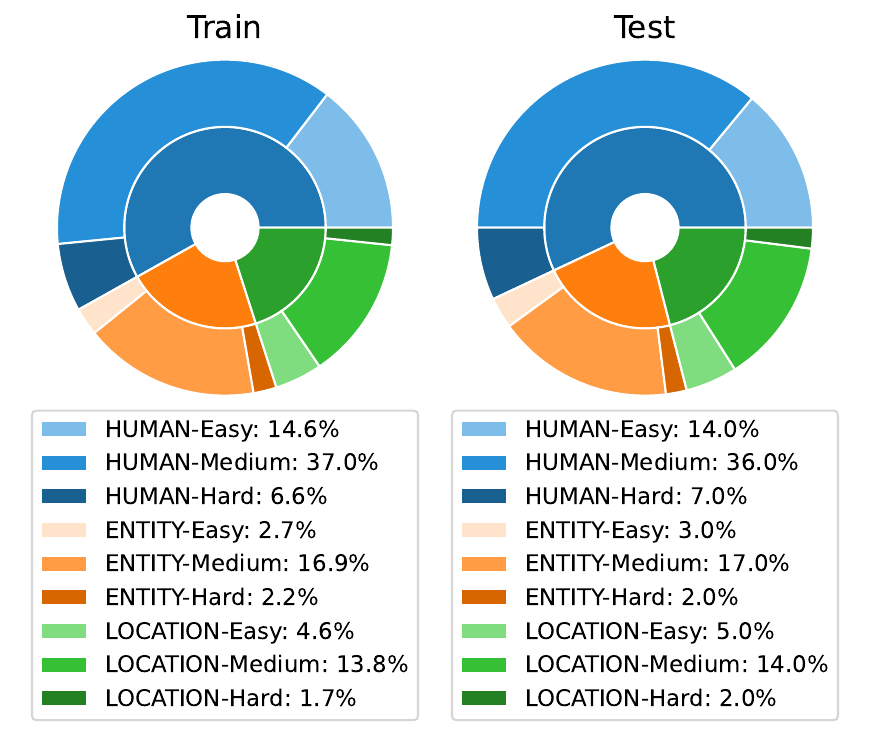}
	\caption{Question difficulty based on the question types.}
	\label{fig:difficulty}
        \Description{}
\end{figure}

\paragraph{Relevance}
This metric assesses the semantic connection between the hint and the question, ensuring that the hint remains meaningful and contextually appropriate. Hints can be considered a form of an answer since they provide explanations of the correct answer of the question. Based on this, one can evaluate the Relevance of a hint to its question as an Answer Relevance task~\cite{es-etal-2024-ragas} - the task where the goal is to assess how pertinent the provided answer is to the target question. To compute the answer relevance metric, we employ DeepEval framework\footnote{\url{https://docs.confident-ai.com/}} treating the hint as a kind of answer.

\paragraph{Readability}
This metric measures the ease of understanding a hint or question, ensuring clarity and accessibility. It is crucial for making the hints effective without confusing users. To evaluate Readability~\cite{liu-lee-2023-hybrid}, we finetune a RoBERTa~\cite{Liu2019RoBERTaAR} model as a classifier on the OneStopEnglish dataset~\cite{vajjala-lucic-2018-onestopenglish}. The finetuned model categorizes sentences into three classes: Beginner (0), Intermediate (1), and Advanced (2), reflecting their level of reading difficulty\footnote{The accuracy of the readability estimator model is 62.3\%}.

\paragraph{Convergence}
This metric measures how effectively a hint can narrow down or eliminate potential answers to the given question. To evaluate hints based on Convergence, we adopt the approach proposed by~\citet{triviahg_mozafari}. In this method, candidate answers are first generated for a question. Then, the metric determines how many candidate answers are supported by the target hint. The convergence score is then derived on the basis of the capability of the hint to eliminate many incorrect candidate answers. We utilize two core models for convergence evaluation: LLaMA-3.1-8B and LLaMA-3.1-70B~\cite{2024arXiv240721783D}.

\paragraph{Familiarity}
This metric measures the degree to which users are expected to know the information expressed in the hints. To evaluate hints based on Familiarity, we follow the approach proposed by~\citet{triviahg_mozafari}. In this method, the named entities are first extracted from the evaluated hint, and the number of views for their corresponding Wikipedia pages is retrieved. After normalizing these values, the final familiarity score is computed as the average of the normalized view counts.

\begin{table}
	\centering
	\small
    \caption{Statistics of \datasetname dataset.}
	\begin{tabular}{@{}l@{\hspace{100pt}}ll@{}}
		\toprule
		                             & Train & Test  \\ \midrule
            Number of hints              & 4,500  & 500   \\		
            Number of questions          & 900   & 100   \\ \midrule
		Avg. question length (words) & 19.55 & 19.19 \\
		Avg. hint length (words)     & 17.77 & 18.32 \\
		Avg. \#entities / question   & 1.2   & 1.44  \\
		Avg. \#entities / hint       & 1.2   & 1.18  \\ \bottomrule
	\end{tabular}
	\label{tbl:statistics}
\end{table}

\begin{table}
	\centering
    \small
    \caption{Question distribution by source and difficulty.}
    \begin{tabular}{@{}l@{\hspace{50pt}}|ccc@{}}
    \toprule
    Source      & ChatGPT & Natural Questions   & SQuAD 2.0 \\ \midrule
    Difficulty & 0.43    & 0.34 & 0.38      \\ \bottomrule
    \end{tabular}
	\label{tbl:dataset_distribution}
\end{table}

\paragraph{Answer Leakage}
The last metric measures the degree to which a hint directly reveals the answer. This metric ensures that hints guide users toward the solution without explicitly disclosing it, promoting problem-solving rather than direct answer retrieval. To compute the degree of Answer Leakage, we assess the semantic similarity between each word in a hint and the corresponding answer using the RoBERTa model. The final score is obtained by maximizing and averaging these similarity values.

\subsection{\evaluatorname Method}\label{ss:hintrank_method}
In addition to the above automatic evaluation approaches involving individual hints, we introduce a new lightweight evaluation method, \evaluatorname, for evaluating and ranking hints using pairwise preferences. Building on the success of widely-used automatic evaluation metrics like BERTScore~\cite{zhang2019bertscore}, BEM~\cite{bulian-etal-2022-tomayto}, MoverScore~\cite{zhao-etal-2019-moverscore}, and BLEURT~\cite{sellam-etal-2020-bleurt}, which leverage BERT~\cite{devlin-etal-2019-bert} as the core evaluation module and demonstrate its effectiveness, we chose BERT as the base for the \evaluatorname method. Our method determines the better hint within a pair of hints. Figure~\ref{fig:evaluation_method} illustrates the proposed method.
In \evaluatorname, we begin by concatenating a given question and its answer with two hints, labeled as $\text{Hint}_{1}$ and $\text{Hint}_{2}$ to create an input compatible with the BERT model. Note that in the answer-agnostic scenarios, we avoid appending the answer to the evaluated hints. Such constructed input is then processed by BERT model, which produces one of two possible outputs: 0 or 1. An output of 0 means that $\text{Hint}_{2}$ is of higher quality than $\text{Hint}_{1}$, whereas an output of 1 suggests that $\text{Hint}_{1}$ is superior to $\text{Hint}_{2}$. As \evaluatorname operates on pairwise preferences, it requires $\binom{n}{2}$ comparisons for a question with $n$ hints, with a runtime complexity of $\mathcal{O}(n^2)$. 

\begin{table*}
\small
\centering
    \caption{Quality comparison of \datasetname and TriviaHG. Relevance, convergence, familiarity, and answer leakage are measured on a scale from 0 to 1, while readability is rated on a scale from 0 to 2 (the lower, the more readable).}
    \begin{tabular}{@{}ll@{\hspace{20pt}}|@{\hspace{20pt}}ccccccc@{}}
\toprule
Dataset     & Subset & \multicolumn{1}{l}{Relevance} & \multicolumn{1}{l}{Readability} & \multicolumn{1}{l}{Convergence} & \multicolumn{1}{l}{Familiarity} & \multicolumn{1}{l}{Length} & \begin{tabular}[c]{@{}c@{}}Answer Leakage Degree\\ (Average)\end{tabular} & \begin{tabular}[c]{@{}c@{}}Answer Leakage Degree\\(Maximum)\end{tabular} \\ \midrule
TriviaHG    & Entire & 0.95                           & \textbf{0.71}                            & 0.57                            & \textbf{0.77}                            & 20.82                      & \textbf{0.23}                                                                  & \textbf{0.44}                                                                  \\
\datasetname & Entire & \textbf{0.98}                           & 0.72                            & \textbf{0.73}                            & 0.75                            & \textbf{17.82}                      & 0.24                                                                  & 0.49                                                                  \\ \midrule \midrule
TriviaHG    & Train  & 0.95                           & 0.73                            & 0.57                            & 0.75                            & 21.19                      & \textbf{0.22}                                                                  & \textbf{0.44}                                                                  \\
\datasetname & Train  & \textbf{0.98}                           & \textbf{0.71}                            & \textbf{0.74}                            & \textbf{0.76}                            & \textbf{17.77}                      & 0.24                                                                  & 0.49                                                                  \\ \midrule
TriviaHG    & Test   & 0.95                           & \textbf{0.73}                            & 0.6                             & \textbf{0.77}                            & 20.97                      & \textbf{0.23}                                                                  & \textbf{0.44}                                                                  \\
\datasetname & Test   & \textbf{0.98}                           & 0.83                            & \textbf{0.72}                            & 0.73                            & \textbf{18.32}                      & 0.24                                                                  & 0.47                                                                  \\ \bottomrule
\end{tabular}
    \label{tbl:quality_wiki_triviahg}
\end{table*}

\section{Experiments and Results} \label{s:experiments_results}
In this section, we first analyze the \datasetname dataset from several different perspectives. Following that, we present the results of the human evaluation. We then examine the performance of different LLMs on the dataset. Lastly, we compare the \evaluatorname method and the convergence metric in different scenarios.
\subsection{Data Analysis} \label{ss:data_analysis}

The \datasetname dataset is split into a train set with 4,500 hints (900 questions) and a test set with 500 hints (100 questions). Table~\ref{tbl:statistics} provides the statistics of the train and test sets, while Figure~\ref{fig:dataset_statistics} shows their distributions according to the question types, indicating that the distributions are well matched.

\begin{figure}
	\centering
	\includegraphics[width=0.80\columnwidth]{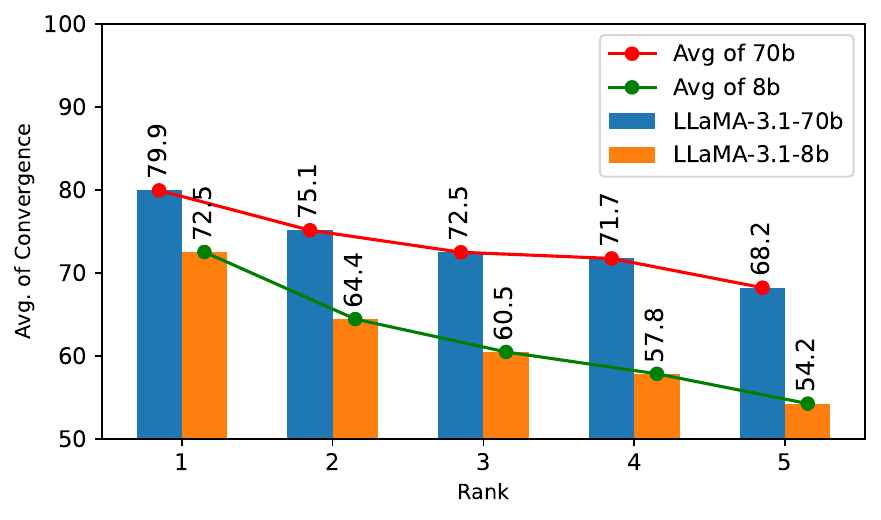}
	\caption{Average convergence of the hints of \datasetname based on the hint ranks.}
	\label{fig:dataset_convergence_rank}
        \Description{}
\end{figure}

We next analyze the difficulty levels of questions in \datasetname. To evaluate the difficulty, we utilize the Reference-based Question Complexity method~\cite{Gabburo2024}. This method computes the difficulty of a question by assessing how many of its retrieved passages contain the correct answer and by measuring the relevance between the retrieved passages and the question. It then calculates the difficulty score for the question based on such computed features. We use the DPR method~\cite{karpukhin-etal-2020-dense} as the retrieval technique, employing the English Wikipedia dump preprocessed by~\citet{karpukhin-etal-2020-dense} as the evidence source, and consider the top 30 most relevant passages as the retrieved passages. Figure~\ref{fig:difficulty} illustrates the computed question difficulty of \datasetname for the train and test subsets\footnote{We classify questions with difficulty scores below 0.33 as easy, those above 0.66 as hard and the rest as medium.}. The figure indicates that medium-hard questions are the most common as well as the train and test subsets have quite similar distributions in terms of question difficulty. Also, Table~\ref{tbl:dataset_distribution} highlights the difficulty levels of questions generated or extracted from various sources.

\begin{table}
\centering
\small
\caption{Demographic information of evaluators}
\begin{tabular}{@{}c@{\hspace{35pt}}c@{\hspace{35pt}}c@{\hspace{35pt}}c@{}}
\toprule
\textbf{Evaluator} & \textbf{Gender} & \textbf{Age} & \textbf{Education Level} \\ \midrule
1      & Female & 18  & High School \\
2      & Female & 36  & High School \\
3      & Female & 30  & Bachelor's Degree \\
4      & Male   & 40  & Master's Degree  \\
5      & Male   & 33  & PhD  \\ \bottomrule
\end{tabular}
\label{tbl:evaluator_background}
\end{table}

\begin{table}
    \centering
    \small
    \caption{Average length of hints vs. their ranks.}
    \begin{tabular}{@{}l@{\hspace{55pt}}|lllll@{}}
        \toprule
        Rank              & 1     & 2     & 3     & 4     & 5    \\ \midrule
        Average Length & 16.99 & 17.67 & 18.02 & 18.14 & 18.3 \\ \bottomrule
    \end{tabular}
	\label{tbl:dataset_length_rank}
\end{table}

Table~\ref{tbl:dataset_length_rank} reveals an interesting insight regarding the length of hints, which can be considered as one of indicators of helpfulness. The results suggest that \textit{high-quality hints tend to be shorter in length (measured by the number of words) than the lower-quality hints.} This finding indicates an inverse correlation between the hint length and helpfulness, challenging the intuition that longer hints are more informative or specific, and therefore more useful. In contrast, shorter hints appear to be more concise and easier to follow, likely presenting more helpful information in the first place.

\begin{table*}
	\caption{Evaluation of generated hints based on relevance, readability, convergence, familiarity, length, and answer leakage across different scenarios.}
	\label{tbl:model_performance}
	\resizebox{\linewidth}{!}{%
	\begin{tabular}{@{}llccccccccc@{}}
\toprule
Model          & Config & \begin{tabular}[c]{@{}c@{}}Use \\ Answer\end{tabular} & \multicolumn{1}{l}{Relevance} & \multicolumn{1}{l}{Readability} & \begin{tabular}[c]{@{}c@{}}Convergence\\ (LLaMA 8b)\end{tabular} & \begin{tabular}[c]{@{}c@{}}Convergence\\ (LLaMA 70b)\end{tabular} & \multicolumn{1}{l}{Familiarity} & \multicolumn{1}{l}{Length} & \begin{tabular}[c]{@{}c@{}}Answer Leakage \\ Degree (Avg)\end{tabular} & \begin{tabular}[c]{@{}c@{}}Answer Leakage\\ Degree (Max)\end{tabular} \\ \midrule
GPT-4          & Vanilla       & \checkmark       & 0.91                          & \textbf{1.0}                               & \textbf{0.14}                                                             & \textbf{0.48}                                                              & 0.84                            & \textbf{26.36}                      & 0.23                                                                   & 0.51                                                                  \\
GPT-4          & Vanilla       & \ding{55}      & 0.92                          & 1.1                             & 0.12                                                             & 0.47                                                              & 0.81                            & 26.93                      & 0.24                                                                   & 0.52                                                                  \\ \midrule
LLaMA-3.1-405b & Vanilla       & \checkmark       & \textbf{0.94}                             & 1.49                            & 0.11                                                             & 0.47                                                              & 0.76                            & 41.81                      & 0.23                                                                   & 0.5                                                                   \\
LLaMA-3.1-405b & Vanilla       & \ding{55}      & 0.92                          & 1.53                            & 0.1                                                              & 0.45                                                              & 0.78                            & 50.91                      & 0.23                                                                   & 0.5                                                                   \\ \midrule
LLaMA-3.1-70b  & FTwA          & \checkmark       & 0.88                          & 1.5                             & 0.09                                                             & 0.42                                                              & \textbf{0.84}                            & 43.69                      & \textbf{0.22}                                                                   & \textbf{0.48}                                                                  \\
LLaMA-3.1-70b  & Vanilla       & \checkmark       & 0.86                             & 1.53                            & 0.05                                                             & 0.42                                                              & 0.8                             & 45.51                      & 0.23                                                                   & 0.5                                                                   \\
LLaMA-3.1-70b  & FTwoA         & \ding{55}      & 0.86                          & 1.5                             & 0.08                                                             & 0.38                                                              & 0.8                             & 51.07                      & 0.22                                                                   & 0.51                                                                  \\
LLaMA-3.1-70b  & Vanilla       & \ding{55}      & 0.87                             & 1.56                            & 0.06                                                             & 0.38                                                              & 0.76                            & 53.24                      & 0.22                                                                   & 0.5                                                                   \\ \midrule
LLaMA-3.1-8b   & FTwA          & \checkmark       & 0.78                          & 1.63                            & 0.05                                                             & 0.37                                                              & 0.79                            & 50.33                      & 0.22                                                                   & 0.52                                                                  \\
LLaMA-3.1-8b   & Vanilla       & \checkmark       & 0.81                          & 1.72                            & 0.05                                                             & 0.32                                                              & 0.8                             & 54.38                      & 0.22                                                                   & 0.5                                                                   \\
LLaMA-3.1-8b   & FTwoA         & \ding{55}      & 0.76                          & 1.7                             & 0.03                                                             & 0.32                                                              & 0.8                             & 55.02                      & 0.22                                                                   & 0.51                                                                  \\
LLaMA-3.1-8b   & Vanilla       & \ding{55}      & 0.78                          & 1.76                            & 0.04                                                             & 0.3                                                               & 0.83                            & 52.99                      & 0.22                                                                   & 0.5                                                                   \\ \bottomrule
\end{tabular}%
	}
\end{table*}

We also evaluate the hints in the entire \datasetname dataset (and separately, in its train and test subsets) using the relevance, readability, convergence, familiarity, length, and answer leakage degree. We then compare these values with the ones obtained for the TriviaHG dataset~\cite{triviahg_mozafari} - the highest-quality and largest hint dataset available. The comparison results are presented in Table~\ref{tbl:quality_wiki_triviahg}. The results indicate that in terms of relevance, readability, answer leakage degree, and familiarity, the metrics are nearly the same between the two datasets. However, \datasetname has better convergence values compared to TriviaHG. Additionally, the hints in \datasetname are shorter, as measured by the word count. These results indicate that the hints in \datasetname are of higher quality. 

Lastly, Figure~\ref{fig:dataset_convergence_rank} demonstrates the correlation between the convergence scores of hints and their helpfulness as represented by hint ranks assigned by crowdworkers.
The plot suggests that \textit{the convergence scores can be considered a reliable metric for evaluating the helpfulness of hints and can be used for hint ranking.} In other words, the higher the convergence, the higher the rank of the hint.

\subsection{Human Evaluation} \label{ss:human_evaluation}
To manually evaluate hints, we recruited five independent evaluators, whose backgrounds are shown in Table~\ref{tbl:evaluator_background}. The evaluators were not involved in the dataset generation process and were tasked with answering questions from the test subset of \datasetname, which consists of 100 questions and 500 hints. The evaluation process followed these steps:
\begin{enumerate}
    \item Participants were asked to answer the question without using any hints. If they provided a correct answer, they proceeded to the next question.
    \item If they could not answer the question correctly, they were asked to review the hints until they could find the correct answer. By providing the correct answer, the participants could move to the next question.
    \item If the participants could not answer the question after reviewing all the hints, they were allowed to skip the question.
\end{enumerate}

\begin{figure}
	\centering
	\includegraphics[width=0.8\columnwidth]{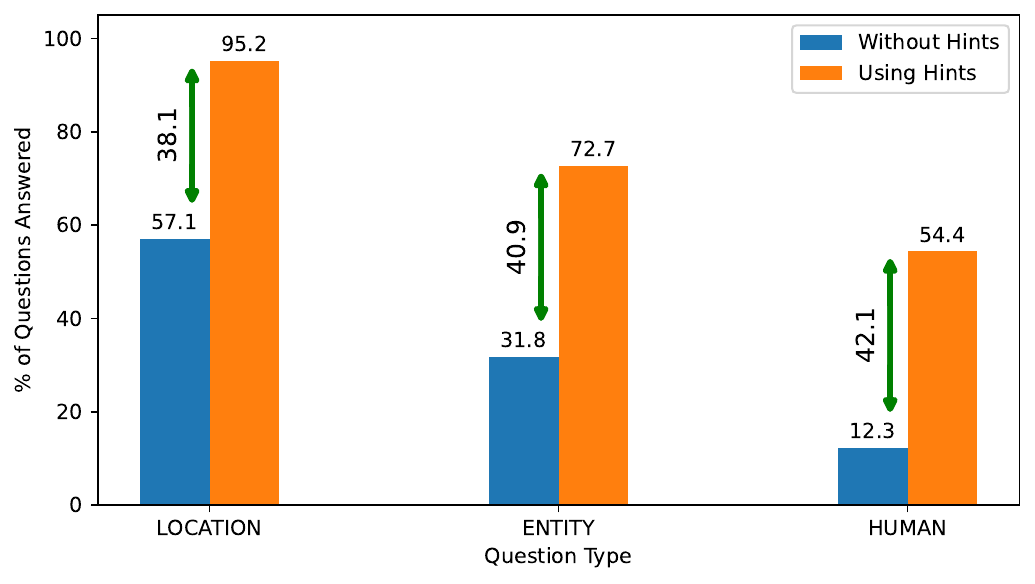}
	\caption{The results of human evaluation.}
	\label{fig:human_evaluation}
        \Description{}
\end{figure}

Figure~\ref{fig:human_evaluation} illustrates that all the participants could answer more questions in question types such as HUMAN, ENTITY, and LOCATION\footnote{We use names as stated in the original dataset.} when they used hints compared to the case without hints. Notably, the greatest improvement was observed in human-related questions, where hints proved most beneficial. Following, entity-related questions led to significant improvement, while location-related questions saw the smallest positive change. This suggests that generating effective hints becomes progressively more challenging for human, entity, and location questions, in that order.

\begin{figure}
	\centering
	\includegraphics[width=0.9\columnwidth]{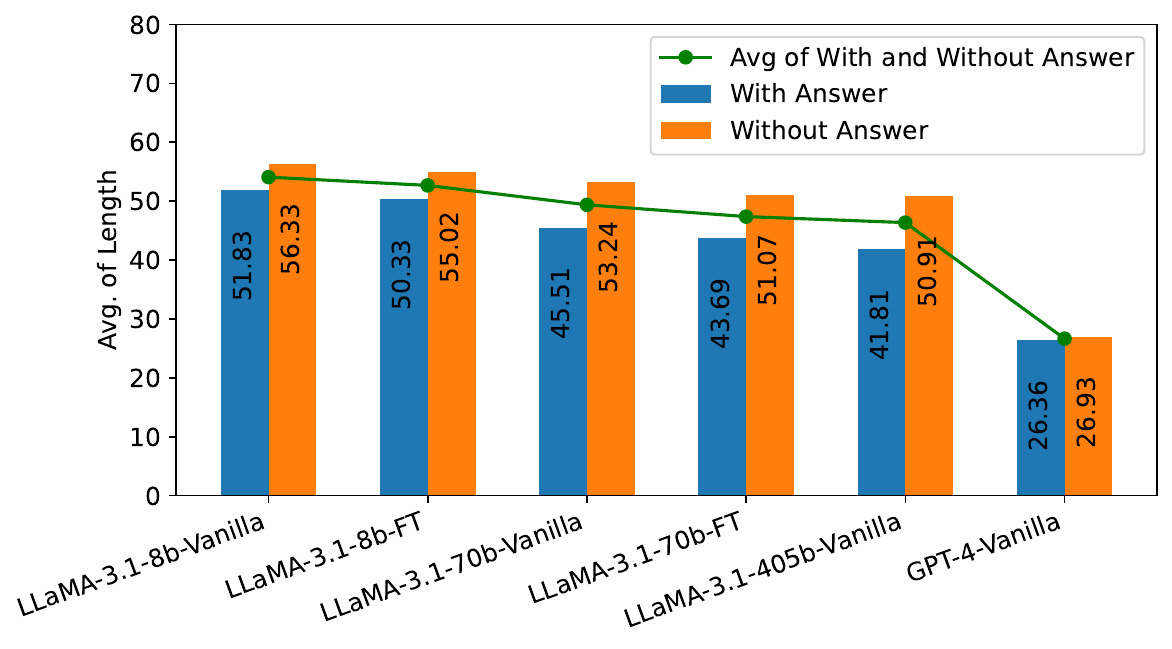}
	\caption{Average length of hints generated by LLMs.}
	\label{fig:model_length_rank}
        \Description{}
\end{figure}

\subsection{Model Performance} \label{ss:model_performance}

To further assess the quality of hints in our dataset, we analyze how well LLMs can automatically generate hints for input questions. We use the open-source LLaMA models: LLaMA-3.1-8b, LLaMA-3.1-70b, and LLaMA-3.1-405b~\cite{2024arXiv240721783D}, and GPT-4~\cite{2023arXiv230308774O}.

\begin{figure*}
	\centering
	\includegraphics[width=0.9\linewidth]{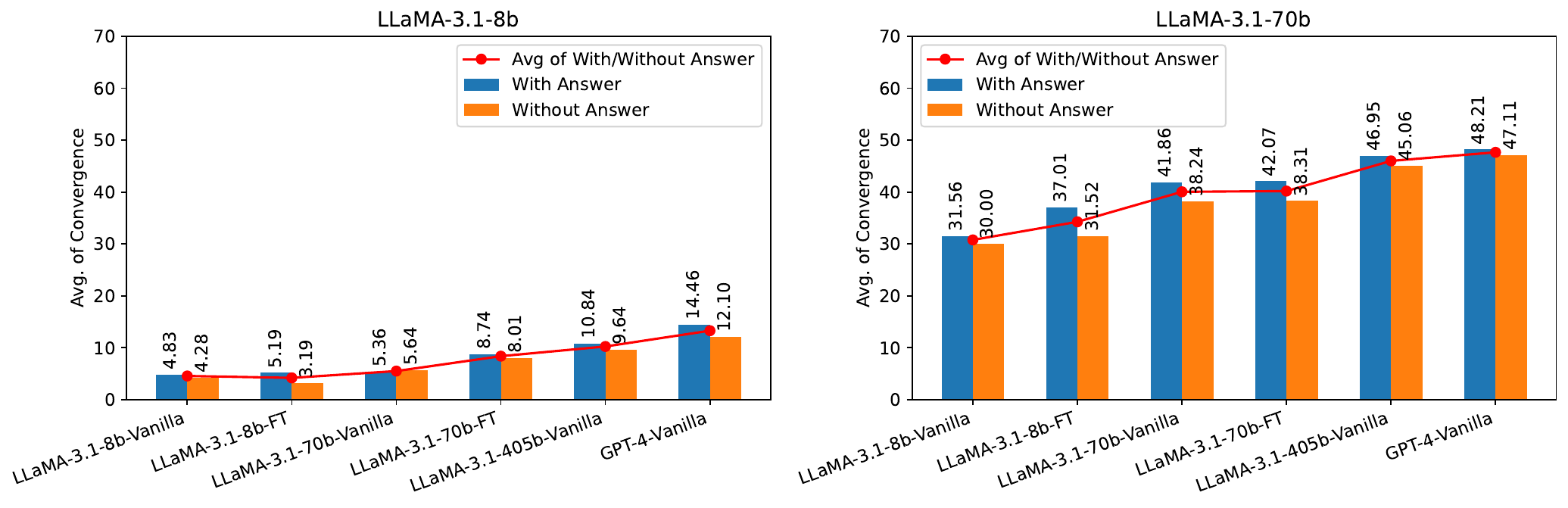}
	\caption{Average convergence of the generated hints by different LLMs. The order of LLMs is determined by their capabilities and the parameter count.}
	\label{fig:models_convergence}
        \Description{}
\end{figure*}

To explore different scenarios, we finetune\footnote{We perform model finetuning using the API functions available on together.ai.} LLaMA-3.1-8b and LLaMA-3.1-70b on the train subset of the \datasetname dataset to evaluate the LLMs' capabilities in hint generation when trained specifically on this task. For each question, we assign a hint as the target during the finetuning process. As a result of this learning strategy, during the inference stage, the finetuned model is prompted to generate one hint for each question.

\begin{table}
	\caption{Comparison between Convergence metric, LLM-based ranking, and \evaluatorname.}
	\label{tbl:conv_pairwiseranker}
	\resizebox{\columnwidth}{!}{%
		\begin{tabular}{@{}llccc@{}}
			\toprule
			Method        & Config & Use Answer & Accuracy (\%) & Correlation (\%) \\ \midrule
			Convergence   & Vanilla       & \checkmark & 40.80     & 36.70        \\ \midrule
			LLaMA-3.1-8b  & Vanilla       & \ding{55}  & 60.50     & 49.25       \\
			LLaMA-3.1-8b  & Vanilla       & \checkmark & 60.95    & 49.79       \\
			LLaMA-3.1-8b  & FTwoA         & \ding{55}  & 61.00       & 50.74       \\
			LLaMA-3.1-8b  & FTwA          & \checkmark & 61.25    & 49.03       \\
			LLaMA-3.1-70b & Vanilla       & \ding{55}  & 64.00       & 50.32       \\
			LLaMA-3.1-70b & Vanilla       & \checkmark & 64.25    & 51.32       \\
			LLaMA-3.1-70b & FTwoA         & \ding{55}  & 64.65    & 51.51       \\
			LLaMA-3.1-70b & FTwA          & \checkmark & 65.30     & \textbf{52.53}       \\ \midrule
			\evaluatorname          & FTwoA         & \ding{55}  & 67.25    & 49.06       \\
			\evaluatorname          & FTwA          & \checkmark & \textbf{68.55}    & 52.34       \\ \bottomrule
		\end{tabular}%
	}
\end{table}

We consider two finetuning approaches depending on whether the answer is or is not used in the input to LLMs: \texttt{Answer-Aware} and \texttt{Answer-Agnostic}. Given that LLMs typically handle most knowledge questions correctly, the answer-agnostic approach might be sufficient for generating hints. Besides, users generally do not know the answers to the questions for which they need hints. However, the answer-aware approach still has its own use, such as in educational contexts where a teacher might wish to collect materials for class preparation. Due to the importance of both approaches, we chose to investigate fine-tuning of the LLMs in these two distinct scenarios.
We found that shorter prompts were more effective in achieving the desired task. Longer, more detailed instructions often led to the model disregarding the key goal of generating hints, and instead focusing on irrelevant details. In contrast, shorter prompts increased the likelihood of successful task completion. After experimenting, we opted for the following prompt as the system prompt:

\begin{center}
\fcolorbox{black!75!black}{framegray}{%
  \begin{minipage}{0.9\linewidth}
    \footnotesize
    \emph{You are a hint generator for the factoid questions. The user asks you a question and you should generate a hint for that question without revealing the answer in the hint.}
  \end{minipage}
}
\end{center}

Two distinct user prompts were employed to generate hints within a zero-shot learning strategy. Assuming a question $q$ as an input, the answer-agnostic prompt was `\emph{Give me the best hint for this question: $q$'}. The answer-aware prompt included the answer $a$ as follows: `\emph{Give me the best hint for this question: $q$? The answer to the question is $a$}'.

To evaluate the hint generation capabilities of LLMs across different scenarios, we examine four approaches including \texttt{Vanilla-wA}, \texttt{Vanilla-woA}, \texttt{FTwA}, and \texttt{FTwoA}, where \texttt{FT} stands for Finetuned, \texttt{wA} denotes With Answer, and \texttt{woA} means Without Answer. We test these models on the \datasetname test set to analyze the impact of finetuning, the inclusion of answers in the prompt, and their overall effectiveness in generating informative hints.

Figure~\ref{fig:model_length_rank} illustrates that as LLMs increase in their size and hint generation capability, the length of the generated hints decreases. This supports our observation made in Section~\ref{ss:data_analysis} of an inverse correlation between hint length and hint quality. Additionally, hints produced by finetuned models are generally shorter than those from vanilla models, indicating that finetuned models may generate higher-quality hints. Moreover, hints in the answer-aware scenarios are shorter compared to those in answer-agnostic scenarios, suggesting that \textit{when the answer is provided along with the question, LLMs are able to produce more effective hints.}

\begin{figure*}
	\centering
	\includegraphics[width=\linewidth]{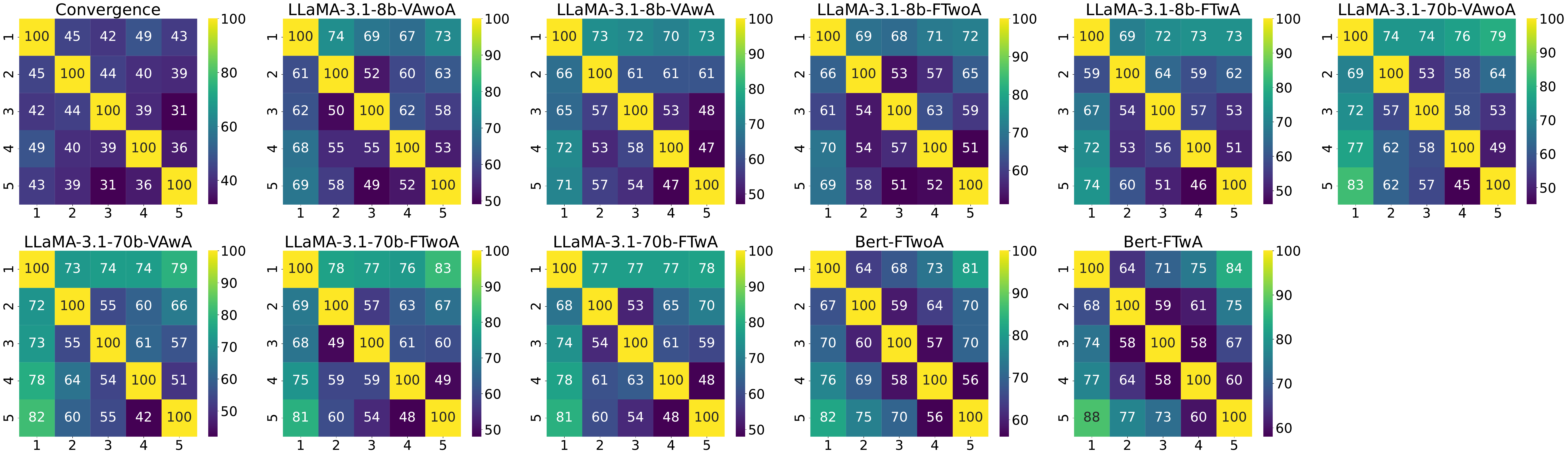}
	\caption{Accuracy of \evaluatorname for different hint pairs in different scenarios. The element at position $(r, c)$ represents the accuracy when comparing $\text{Hint}_1$ at rank $r$ to $\text{Hint}_2$ at rank $c$.}
	\label{fig:evaluation_correlation_matrix}
        \Description{}
\end{figure*}

Table~\ref{tbl:model_performance} presents the quality of generated hints, evaluated with methods such as relevance, readability, convergence, answer leakage degree, and familiarity. 
The results indicate that more powerful LLMs are capable of generating more relevant hints.
Regarding readability, GPT-4 exhibits the highest quality, followed closely by LLaMA-3.1-405b and LLaMA-3.1-70b, while LLaMA-3.1-8b shows the lowest readability. It also demonstrates that more powerful LLMs can generate more readable hints. Additionally, finetuned models consistently outperform their vanilla counterparts, and answer-aware prompts yield better results compared to answer-agnostic prompts for readability and familiarity. The answer leakage degree indicates that the prompt we use is effective in preventing LLMs from directly including answers, their synonyms or very similar terms in the generated hints, as the results closely align with those of \datasetname shown in Table~\ref{tbl:quality_wiki_triviahg}.

Figure~\ref{fig:models_convergence} illustrates that as LLMs increase in size and capability, the convergence of their hints also increases. This trend is observed for both LLaMA-3.1-8b and LLaMA-3.1-70b, used as the cores of the convergence method. This supports our claim in Section~\ref{ss:data_analysis} regarding the correlation between convergence and ranks. The figure also shows 
that the \emph{average convergence of hints created by the answer-aware approach surpasses that of the ones by answer-agnostic approach, suggesting that including the answer in the prompt makes it easier for LLMs to generate good hints}. Furthermore, LLMs finetuned on the train subset of the \datasetname dataset achieve better convergence scores than their vanilla counterparts, indicating \textit{the usefulness of \datasetname for finetuning LLMs for hint generation.}

\subsection{Evaluation of \evaluatorname Method} \label{ss:evaluation_method}

As outlined in Section~\ref{s:evaluation_method}, we also propose in this paper a novel evaluation method, \evaluatorname, for ranking hints using the BERT model. In addition to finetuning BERT, we additionally finetune LLaMA-3.1-8b and LLaMA-3.1-70b models on the train set of the \datasetname to assess the performance of these LLMs in identifying high-quality hints. Similarly to the experiments described in Section~\ref{ss:model_performance}, we examine various scenarios including answer-aware and answer-agnostic contexts, and compare vanilla models with their finetuned counterparts. We use the following prompt as the system prompt:

\begin{center}
\fcolorbox{black!75!black}{framegray}{%
  \begin{minipage}{0.9\linewidth}
    \footnotesize
    \emph{You are an expert hint evaluator for factoid questions. Given a question and two hints, your task is to determine which hint is more helpful in guiding the user toward the correct answer.}
  \end{minipage}
}
\end{center}

Two distinct user prompts are employed for evaluating hints within a zero-shot learning strategy. Assuming a question $q$ as a question and $h_1$ and $h_2$ as a pair of hints, the answer-agnostic prompt is:

\begin{center}
\fcolorbox{black!75!black}{framegray}{%
  \begin{minipage}{0.9\linewidth}
    \footnotesize
    \emph{Which hint is better to find the answer of this question: $q$. Hint\_1: $h_1$. Hint\_2: $h_2$. Just choose between "Hint\_1" and "Hint\_2" without any explanations.}
  \end{minipage}
}
\end{center}

and the answer-aware prompt with answer $a$ is: 

\begin{center}
\fcolorbox{black!75!black}{framegray}{%
  \begin{minipage}{0.9\linewidth}
    \footnotesize
    \emph{Which hint is better to find the answer of this question: $q$. The answer to this question is $a$. Hint\_1: $h_1$. Hint\_2: $h_2$. Just choose between "Hint\_1" and "Hint\_2" without any explanations.}
  \end{minipage}
}
\end{center}

We benchmark \evaluatorname against the Convergence metric which turned out to be useful for evaluating hint ranking as indicated in Figure~\ref{fig:dataset_convergence_rank}. To convert pairwise rankings to listwise rankings, we apply the Bradley–Terry model~\cite{19ff28b9-64f9-3656-ba40-08326a05748e}. We then evaluate the correlation between the rankings with the Pearson Correlation~\cite{mining2006data}.

Table~\ref{tbl:conv_pairwiseranker} outlines the key features and differences between various scenarios. The results indicate that with the increase in the size and power of LLMs, both accuracy and correlation improve. Additionally, the answer-aware approach yields better outcomes compared to the answer-agnostic method, suggesting that the presence of an answer enables LLMs to evaluate hints more effectively. Moreover, finetuned versions outperform their vanilla counterparts, demonstrating that the \datasetname dataset is well-suited for model fine-tuning to rank hints. Surprisingly, the BERT-base method outperforms LLMs, including their finetuned versions in the answer-aware scenario. This holds true for both \texttt{Bert-FTwoA} and \texttt{Bert-FTwA}, although BERT-base also performs better in the answer-aware approach compared to answer-agnostic. BERT-base methods achieve higher accuracy than LLMs and convergence, but in terms of correlation, \texttt{LLaMA-3.1-70b} exhibits the best performance. The effectiveness of BERT-base methods may be attributed to the strengths of encoder-based models like BERT in classification tasks over decoder-based models. Utilizing BERT-based models instead of LLMs enhances the speed and accessibility of \evaluatorname, reducing computational demands.
Figure~\ref{fig:evaluation_correlation_matrix} shows that the accuracy improves as the rank difference between the hints increases, indicating it is harder to correctly order hints with closer ranks.

\section{Conclusion}
Automatic hint generation is a task closely related to question answering and question generation problems. Effective hints can constitute a useful mechanism for engaging users in the information seeking process.
In this paper, we introduced the first manually created dataset for hint generation and ranking. We also presented a new lightweight method for ranking and evaluating hints. To demonstrate the effectiveness of our dataset, we conducted experiments where humans attempted to answer questions with and without the use of hints. The results confirm that the hints are of sufficient quality to assist users. We then finetuned LLMs using our dataset, prompting them to generate new hints for different questions. The high quality of the generated hints indicates that our dataset is well-suited for finetuning LLMs for the HG task. We also finetuned BERT and LLMs on \datasetname for the task of hint ranking and evaluated their performance. The results reveal that encoder-based models outperform decoder-based models in hint ranking.

In the future, we plan to generate personalized hints tailored to the knowledge of users. The main challenge here will be to develop appropriate datasets and solutions for user profiling.

\bibliographystyle{ACM-Reference-Format}
\balance
\bibliography{Main}

\end{document}